\documentclass{article}
\usepackage{ijcai21}

\usepackage{soul}
\usepackage{url}
\usepackage[hidelinks]{hyperref}
\usepackage[utf8]{inputenc}
\usepackage{booktabs}
\usepackage{times}
\usepackage{epsfig}
\usepackage{graphicx}

\usepackage{subcaption} % comment subfigure package while using it
\usepackage{microtype}
\usepackage{amsmath,amssymb,amsthm}
\usepackage{amsfonts}

\usepackage{enumitem} % to reduce space between items in enumerate

\usepackage{multirow}
\usepackage[table,xcdraw]{xcolor} % for color in the table

\usepackage{algorithm}
\usepackage{algpseudocode}
\algrenewcommand\algorithmicrequire{\textbf{Input:}}
\algrenewcommand\algorithmicensure{\textbf{Output:}}
\usepackage{tabularx}
\makeatletter
\newcommand{\multiline}[1]{%
  \begin{tabularx}{\dimexpr\linewidth-\ALG@thistlm}[t]{@{}X@{}}
    #1
  \end{tabularx}
}
\makeatother

\def\onedot{\ifx\@let@token.\else.\null\fi\xspace}

\makeatother

\def\Vec#1{{\boldsymbol{#1}}} % vectors
\title{Color Variants Identification in Fashion e-commerce via \\Contrastive Self-Supervised Representation Learning}

\author{
Ujjal Kr Dutta
\and
Sandeep Repakula
\and
Maulik Parmar
\And
Abhinav Ravi
\affiliations
Data Sciences-Image Sciences, Myntra
\\Bengaluru, Karnataka, India
\emails
\{ujjal.dutta,sandeep.r,parmar.m,abhinav.ravi\}@myntra.com
}

\begin{document}

\maketitle

\begin{abstract}
In this paper, we utilize deep visual Representation Learning to address an important problem in fashion e-commerce: color variants identification, i.e., identifying fashion products that match exactly in their design (or style), but only to differ in their color. At first we attempt to tackle the problem by obtaining manual annotations (depicting whether two products are color variants), and train a supervised triplet loss based neural network model to learn representations of fashion products. However, for large scale real-world industrial datasets such as addressed in our paper, it is infeasible to obtain annotations for the entire dataset, while capturing all the difficult corner cases. Interestingly, we observed that color variants are essentially manifestations of color jitter based augmentations. Thus, we instead explore Self-Supervised Learning (SSL) to solve this problem. We observed that existing state-of-the-art SSL methods perform poor, for our problem. To address this, we propose a novel SSL based color variants model that simultaneously focuses on different parts of an apparel. Quantitative and qualitative evaluation shows that our method outperforms existing SSL methods, and at times, the supervised model.
\end{abstract}

%%%%%%%%% BODY TEXT
\section{Introduction}
\begin{figure}[t]
  \centering
  \includegraphics[width=0.5\linewidth]{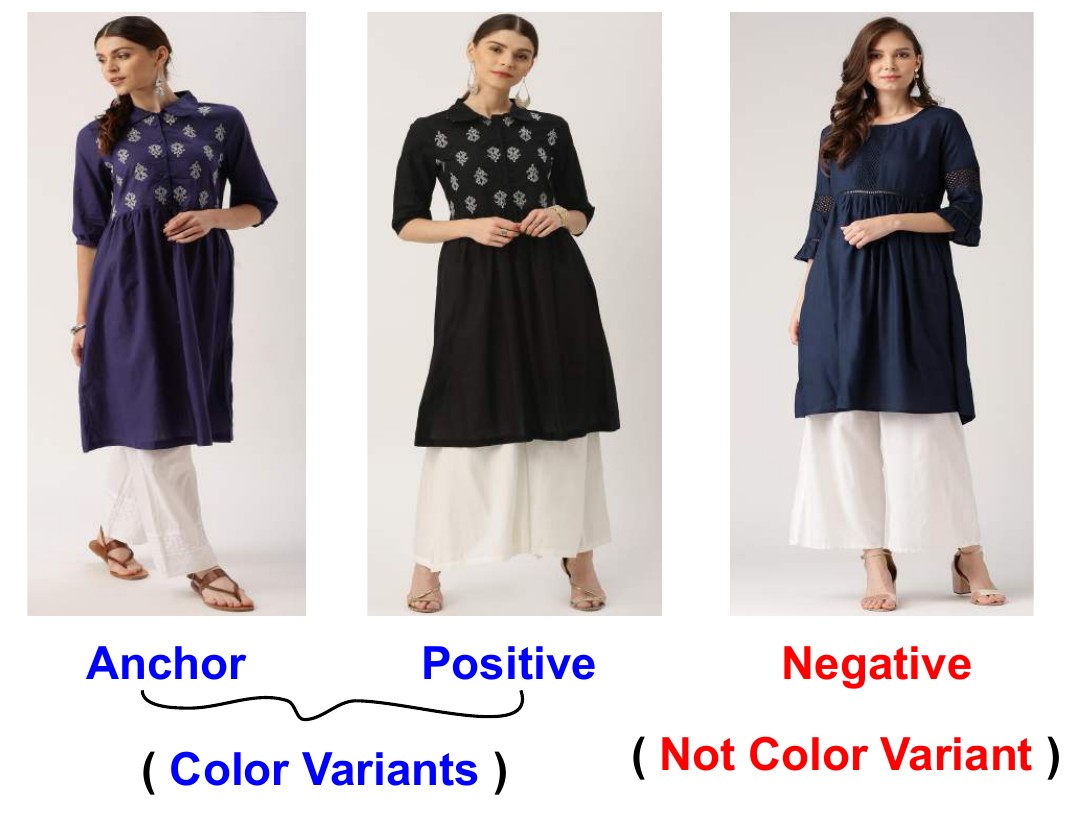}
  \caption{Illustration of the \textit{color variants identification} problem. The anchor and positive images contain fashion products that have the exact same design/ style, but different colors (anchor is blue, and positive is black). The negative image contains a product that is not a color variant to the anchor and the positive. NOTE: The images belong to \url{www.myntra.com}.}
  \label{CV_illustration}
\vspace{-0.7cm}
\end{figure}

In this paper, we address a very crucial problem in fashion e-commerce, namely, automated \textit{color variants identification}, i.e., identifying fashion products that match exactly in their design (or style), but only to differ in their color (Figure \ref{CV_illustration}). Our motivation to pick the use-case of color variants identification for fashion products comes from the following reasons: i) Fashion products top across all categories in online retail sales \cite{jagadeesh2014large}, ii) Most often users hesitate to buy a fashion product solely due to its color despite liking all other aspects of it. Providing more color options increases add-to-cart ratio, thereby generating more revenue, along with improved customer experience. At Myntra (\url{www.myntra.com}), a leading e-commerce platform, we address this problem by leveraging deep visual Representation/Embedding Learning.

Firstly, we obtained manual annotations (class labels) indicating whether two product images are color variants to each other. These class labels are used to obtain triplet based constraints, consisting of an anchor, a positive and a negative (Figure \ref{CV_illustration}). The positive is usually an image that consists of a product that is a color variant of the product contained in the anchor image. The negative is an image that consists of a product that is not a color variant of the products contained in the anchor and positive images. These triplets are used to train a supervised triplet loss based neural network model \cite{schroff2015facenet,veit2017conditional} in order to obtain deep embeddings of fashion products. Having obtained the embeddings, we perform clustering on them to group the color variants.

A key challenge in this supervised approach is that of obtaining manual annotations, which not only requires fashion domain expertise, but is also infeasible, given the large scale of our platform, and the huge number and complexity of the fashion products. As visual Self-Supervised Learning (SSL) obtains image embeddings without requiring manual annotations, we consider it as a candidate to address our problem, i.e., lack of annotations for our large data. The motivation for this comes from the fact that typcial visual SSL methods employ a color jitter based data augmentation step.

\begin{figure}[t]
  \centering
  \includegraphics[width=0.8\linewidth]{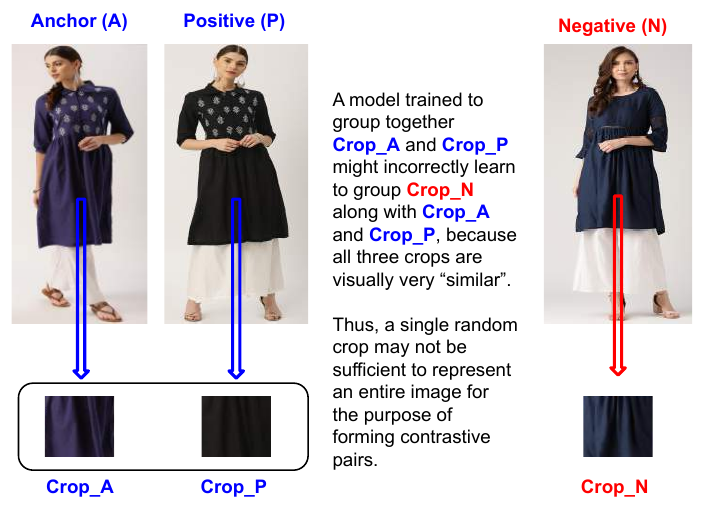}
  \caption{Drawbacks of using random crop to form contrastive pairs in our use-case.}
  \label{slice_motivation}
\vspace{-0.7cm}
\end{figure}

Interestingly, color variants of fashion products are in essence, manifestations of color jittering. It should be noted that a Product Display Page (PDP) image in a fashion e-commerce platform may contain multiple fashion products. Thus, we must apply an object detector to localise our primary fashion product of interest (Figure \ref{CV_pipeline}). However, when we already employ object detection, the standard random crop step used in existing SSL methods may actually miss out important regions of a fashion product, which might be crucial in identifying color variants (Figure \ref{slice_motivation}).

For this reason, we rather choose to propose a novel SSL variation that considers multiple slices/ patches of the primary fashion object (after object detection), and simultaneously obtains embeddings for each of them. The final sum-pooled embedding is then used to optimize a SSL based contrastive loss (Figure \ref{PBCNet_illus}). We call our method as \textbf{Patch-Based Contrastive Net (PBCNet)}. We conjecture that considering multiple patches i) do not leave things to chance (as in random crops), and provides a deterministic approach to obtain embeddings, ii) enables us to borrow more information (from other patches) to make a better decision on grouping a pair of similar embeddings. Our conjecture is supported not only by evidence of improved discriminative performances by the consideration of multiple fine patches of images \cite{wang2018learning}, but also by our experimental results, where our method consistently outperforms existing SSL methods on our task, and also the supervised baseline.
% Motivated by the fact that considering multiple fine patches of images often leads to improved discriminative performances \cite{wang2018learning}, we propose a novel contrastive learning method that obtains embeddings by considering multiple patches of an object. To this end, we propose a simple way to partition the object images into fixed horizontal and vertical stripes. The base encoder of the model obtains vector representations of each of these stripes and adds them to obtain the final embedding of the object. Our conjecture is that by focusing on multiple fixed stripes of an image, the model may be better equipped in identifying the actual discriminative regions of an image that decide whether two objects are color variants or not. We evaluate our hypothesis by comparing our proposed model both quantitatively as well as qualitatively against the state-of-the-art self-supervised representation learning methods and our initial supervised model.

Following are the \textbf{major contributions of the paper}:
\begin{enumerate}[noitemsep,nolistsep]
    \item A supervised triplet loss based visual Representation Learning approach to identify color variants among fashion products (to the best of our knowledge addressed for the first time).
    \item A systematic study of existing state-of-the-art SSL methods to solve the proposed problem while alleviating the need for manual annotations.
    \item A novel contrastive loss based SSL method that focuses on parts of an object to identify color variants.
\end{enumerate}

\section{Related Work}

% \textbf{Deep Embedding Learning:}
The problem of visual embedding/ metric learning refers to that of obtaining vector representations/embeddings of images in a way that the embeddings of similar images are grouped together while moving away dissimilar ones. Several approaches have been proposed in the recent literature, which can be categorized as either supervised \cite{circle_CVPR20,class_collapse_2020,gu2020symmetrical} or unsupervised \cite{dutta2020unsupervised,cao2019unsupervised,li2020unsupervised,SUML_AAAI20}. Typically, a key aspect of such approaches is that of providing constraints for optimizing a suitable objective function. Popular forms of constraints in embedding learning are either in the form of pairs, triplets or tuples, that indicate embeddings of which examples need to be brought closer.%\cite{tuplet_margin_ICCV19,multi_sim_CVPR19,SNR_CVPR19,circle_CVPR20,fastAP_CVPR19,proxyNCA_ICCV17,arcface_CVPR19,soft_triple_ICCV19,class_collapse_2020,gu2020symmetrical,chen2020compressed,li2020symmetric,gong2020online}

% \textbf{Visual Self-Supervised Learning:}
Visual SSL \cite{jing2020self} refers to the learning of representations of images without making use of class labels. A popular paradigm of SSL is contrastive learning, that groups together embeddings obtained from augmentations of the same image. Many recent SSL approaches have been proposed that vary in their implementation details (for example, using momentum encoding, memory module, only positive pairs, etc) \cite{simclr_20,moco_cvpr20,byol_20,simsiam_21}.
% However, contrasting only such positive pairs may lead to a model collapse, where one obtains degenerate solutions by trivially optimizing the objective. To avoid this, negative pairs need to be considered as well. In the contrastive loss scenario, negative pairs are formed by considering augmentations obtained from different images in a mini-batch. Usually, the quality of obtained representations depend on the quality and the number of negatives considered. Approaches like MOCOv2 \cite{moco_cvpr20} maintain a separate memory module to facilitate comparisons with a large number of negatives. However, as mining negative examples is not only crucial, but challenging as well, a few recent approaches (eg, BYOL \cite{byol_20} and SimSiam\cite{simsiam_21}) have been proposed that avoid a model collapse while using only positive pairs, by virtue of crafty modifications.

\section{Proposed Approach}
\begin{figure}[t]
  \centering
  \includegraphics[width=0.8\linewidth]{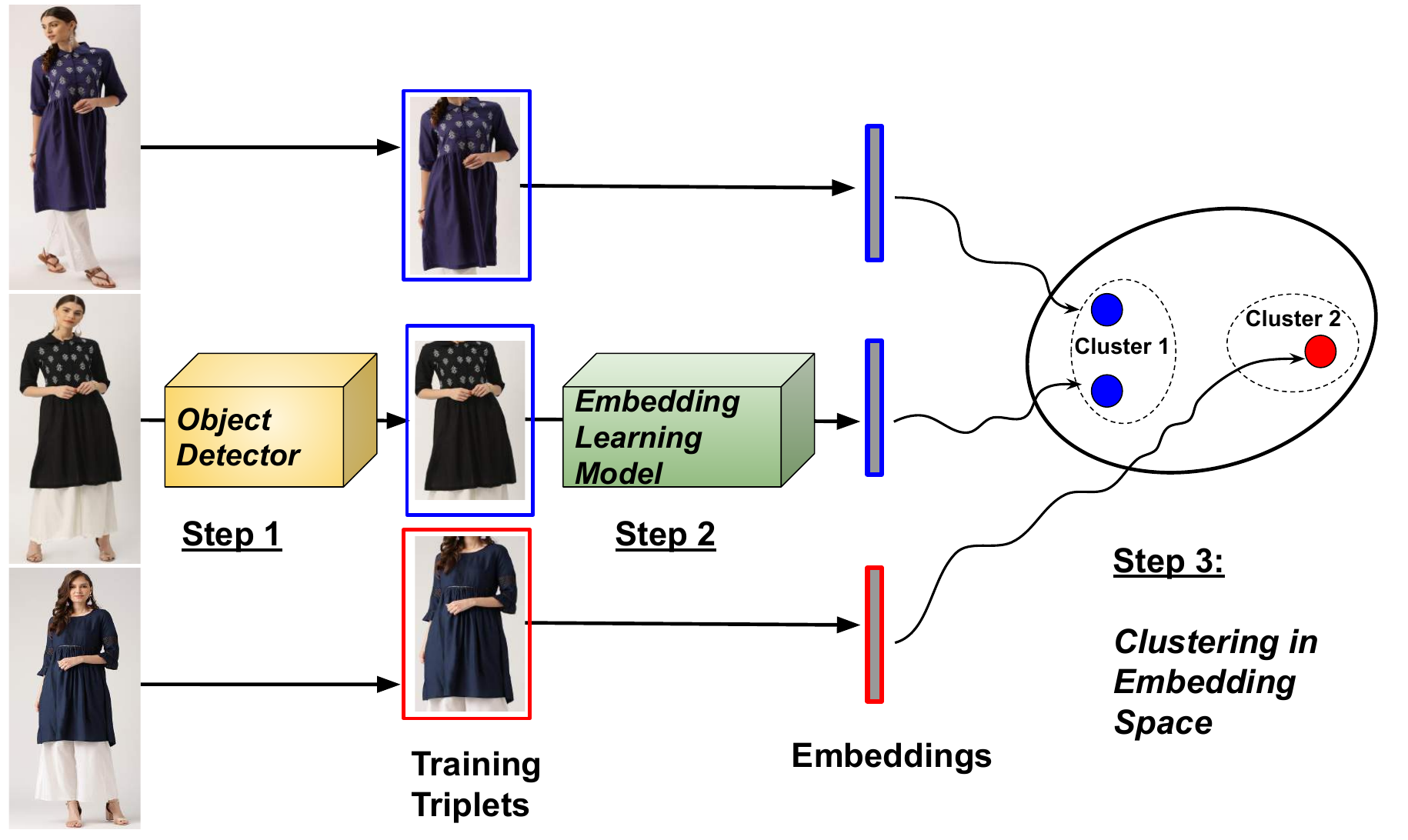}
  \caption{Proposed pipeline to address the color variants identification problem by using a supervised embedding learning model.}
  \label{CV_pipeline}
\vspace{-0.5cm}
\end{figure}
In this section, we discuss the Representation Learning methods used to address the problem of color variants identification.
\subsection{Supervised Color Variants Model}
Firstly, we shall discuss the proposed supervised approach of addressing the color variants identification problem. Our proposed pipeline leveraging the supervised model is illustrated in Figure \ref{CV_pipeline}. The pipeline consists of the following major components (or steps) in the same order: i) Object Detection, ii) Embedding Learning and iii) Clustering. As the original input image usually consists of a human model wearing secondary fashion products as well, we perform object detection to localise the primary fashion product of interest. Having obtained the cropped image of the fashion article, we form triplet based constraints (in the form of anchor, positive and negative) using the available manual annotations. These triplets are used to train an embedding learning model. The obtained embeddings are then grouped together by using an appropriate clustering algorithm. An obtained cluster then contains embeddings of images of fashion products that are color variants to each other.

However, the supervised model needs manual annotations which may be infeasible to obtain in large real-world industrial datasets (such as those present in our platform). Thus, we now propose a novel self-supervised representation learning model to identify color variants without making use of manual annotations.

\subsection{Self-Supervised Color Variants Model}
\begin{figure}[t]
  \centering
  \includegraphics[width=0.85\linewidth]{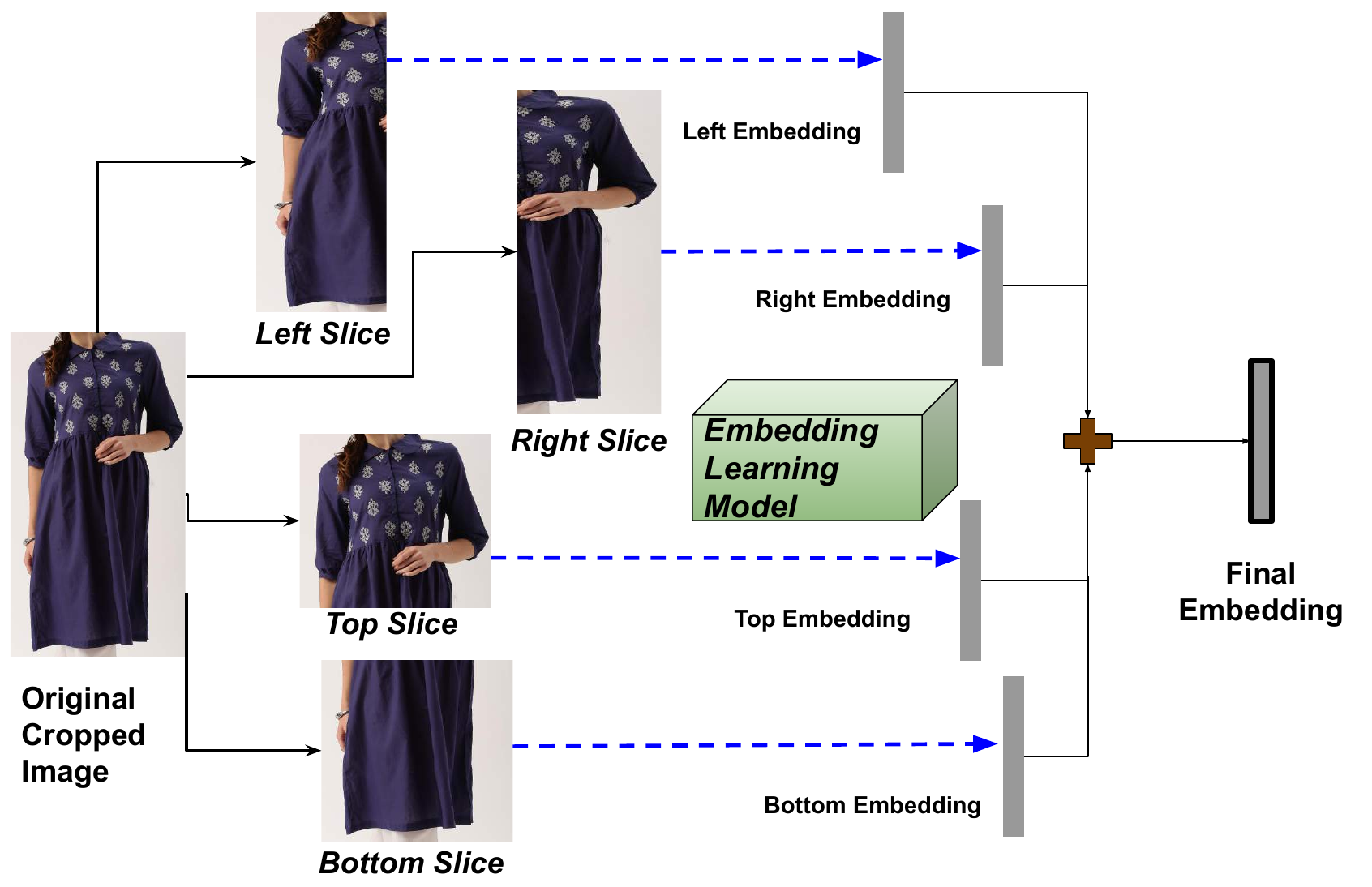}
  \caption{Illustration of our slicing based approach.}
  \label{PBCNet_illus}
\vspace{-0.7cm}
\end{figure}
As discussed using Figure \ref{slice_motivation}, a random crop of an image may not be the best representative to form contrastive pairs for SSL, at least for our task. Hence, we propose a method that simultaneously considers multiple, fixed, slices/patches of an object to form embeddings. This is illustrated in Figure \ref{PBCNet_illus}. As our contribution is only specific to the embedding learning component, Figure \ref{PBCNet_illus} focuses only on this. As illustrated, the multiple patches of an object are obtained by performing slices of the image to obtain the left, right, top and bottom views of the object under consideration. We then pass each of these slices through the base encoder of the embedding learning model to obtain four different embeddings, which are then added\footnote{Rather than concatenating or averaging, we simply add them together to maintain simplicity of the model.} to obtain the final representation of the object.

We make use of negative pairs in our method because we found the performance of methods that do not make use of negative pairs (eg, SimSiam \cite{simsiam_21}, BYOL \cite{byol_20}) to be sub-optimal in our use-case. The objective of our method is similar to the commonly used Normalized Temperature-scaled cross entropy (NT-Xent) loss \cite{simclr_20}. In particular, we make use of two branches for the encoders of our embedding learning model, one for the query and another for the key \cite{simclr_20,byol_20}. In practice, an encoder is a Convolutional Neural Network (CNN) that takes a raw input image and produces an embedding vector. The query encoder is a CNN that obtains embeddings for the anchor images, while the key encoder is a copy of the same CNN that obtains embeddings for the positives and negatives. Gradients are backpropagated only through the query encoder, which is used to obtain the final representations of the inference data.

Following is the objective of our method:
\begin{equation}
    \label{NTXent}
    \mathcal{L}_{\Vec{q}} = - \textrm{log} \frac{ \textrm{exp}( {\Vec{q}\Vec{k}_q} / \tau ) }{ \sum_{i=0}^K \textrm{exp}( {\Vec{q}\Vec{k}_i} / \tau ) }
\end{equation}
Here, $\Vec{q}=\sum_v \Vec{q}^{(v)}$, $\Vec{k}_i=\sum_v \Vec{k}_i^{(v)}, \forall i$. In (\ref{NTXent}), $\Vec{q}$ and $\Vec{k}_i$ respectively denote the \textit{final} embeddings obtained for a query and a key, which are essentially obtained by adding the embeddings obtained from across all the views, as denoted by the superscript $v$ for $\Vec{q}^{(v)}$ and $\Vec{k}_i^{(v)}$. Also, $\Vec{k}_q$ represents the positive key corresponding to a query $\Vec{q}$, $\tau$ denotes the temperature parameter, whereas $\textrm{exp}()$ and $\textrm{log}()$ respectively denote the exponential and logarithmic functions.

During our experiments, we observed that simple methods like SimSiam \cite{simsiam_21} do not actually perform too well for our use-case. On the contrary, we found benefits of components such as the momentum encoder, as present in BYOL \cite{byol_20} and MOCOv2 \cite{moco_cvpr20}. Also, adding an extra memory module in the form of a queue helps in boosting the performance due to comparisons with a large number of negatives. Thus, $K$ in (\ref{NTXent}) denotes the size of the memory module. Our method is called as \textbf{Patch-Based Contrastive Net (PBCNet)}.

\section{Experiments}
We evaluated the discussed methods on a large (orders of magnitudes of $10^5$) internal collection of challenging real-world industrial images on our Myntra platform (\url{www.myntra.com}) that hosts various fashion products. In this section, we report our results on a collection of roughly\footnote{Company compliance policies prohibit open-sourcing/ revealing exact dataset specifics} 0.13 million Kurtas images from our internal database. We used the exact same set to train the supervised (with labeled training data) and self-supervised methods (without labeled training data) for a fair comparison. For inferencing, we used the entire 0.13 million Kurtas images, which are present in the form of different dataset splits (based on brand, gender, MRP). We refer to our 6 dataset splits as Data 1-6. Details of the data and the performance metrics (CGacc for all splits, ARI, FMS and CScore for splits 4-6, a higher value indicates a better performance) are deferred to the supplementary material.% due to space constraints.

\textbf{Methods Compared}: Following are the methods that we have compared for representation learning:
\begin{enumerate}[noitemsep,nolistsep]
\item \textbf{Triplet Loss based Deep Neural Network} \cite{schroff2015facenet,veit2017conditional}: This is our supervised baseline that is trained using triplet based constraints obtained using the labeled data.
\item \textbf{SimSiam} \cite{simsiam_21}: This is a recently proposed State-Of-The-Art (SOTA) visual SSL method that neither uses negative pairs, nor momentum encoder, nor large batches.
\item \textbf{BYOL} \cite{byol_20}: This is another recently proposed SOTA SSL method that also does not make use of negative pairs, but makes use of batch normalization and momentum encoding. 
\item \textbf{MOCOv2} \cite{moco_cvpr20}: This is a SOTA SSL method that makes use of negative pairs, a momentum encoder, as well as a memory queue.   
\item \textbf{PBCNet}: This is our proposed novel self-supervised method.
\end{enumerate}
We implemented all the methods in PyTorch. For all the compared methods, we fix a ResNet34 \cite{ResNet} as a base encoder with  $224 \times 224$ image resizing, and train all the models for a fixed number of 30 epochs, for a fair comparison. The number of epochs was fixed based on observations on the supervised model, to avoid overfitting. For the purpose of object detection, we made use of YOLOv2 \cite{yolov2}, and for the task of clustering the embeddings, we made use of the Agglomerative Clustering algorithm with Ward's method for merging. In all cases, the 512-dimensional embeddings used for clustering are obtained using the avgpool layer of the trained encoder. A margin of 0.2 has been used in the triplet loss for training the supervised model.

\subsection{Systematic Study of SSL for color variants identification}
\begin{figure}[t]
\centering
    \begin{subfigure}{0.45\linewidth}
        \centering
        \includegraphics[width=\linewidth]{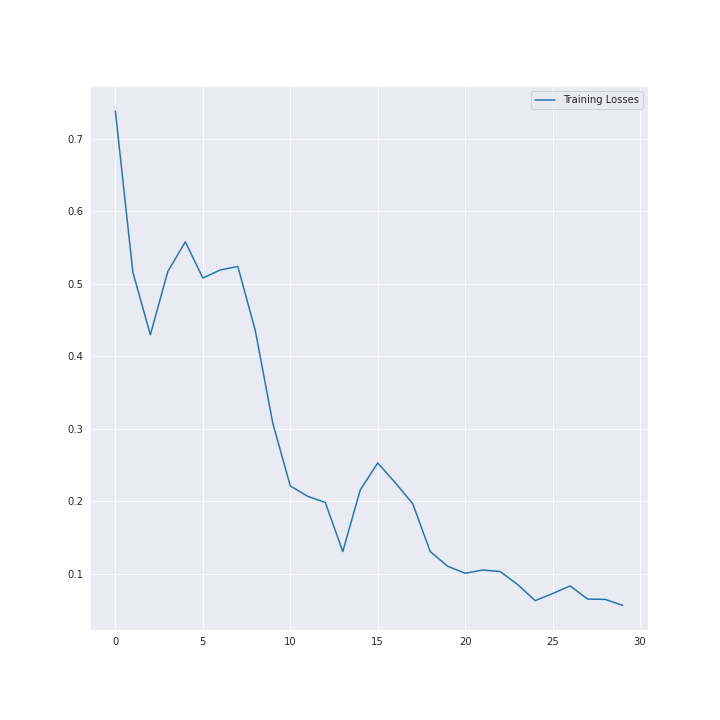}  
        \caption{}
        \label{conv_simsiam_v0}
    \end{subfigure}
    \begin{subfigure}{0.45\linewidth}
        \centering
    	\includegraphics[width=\linewidth]{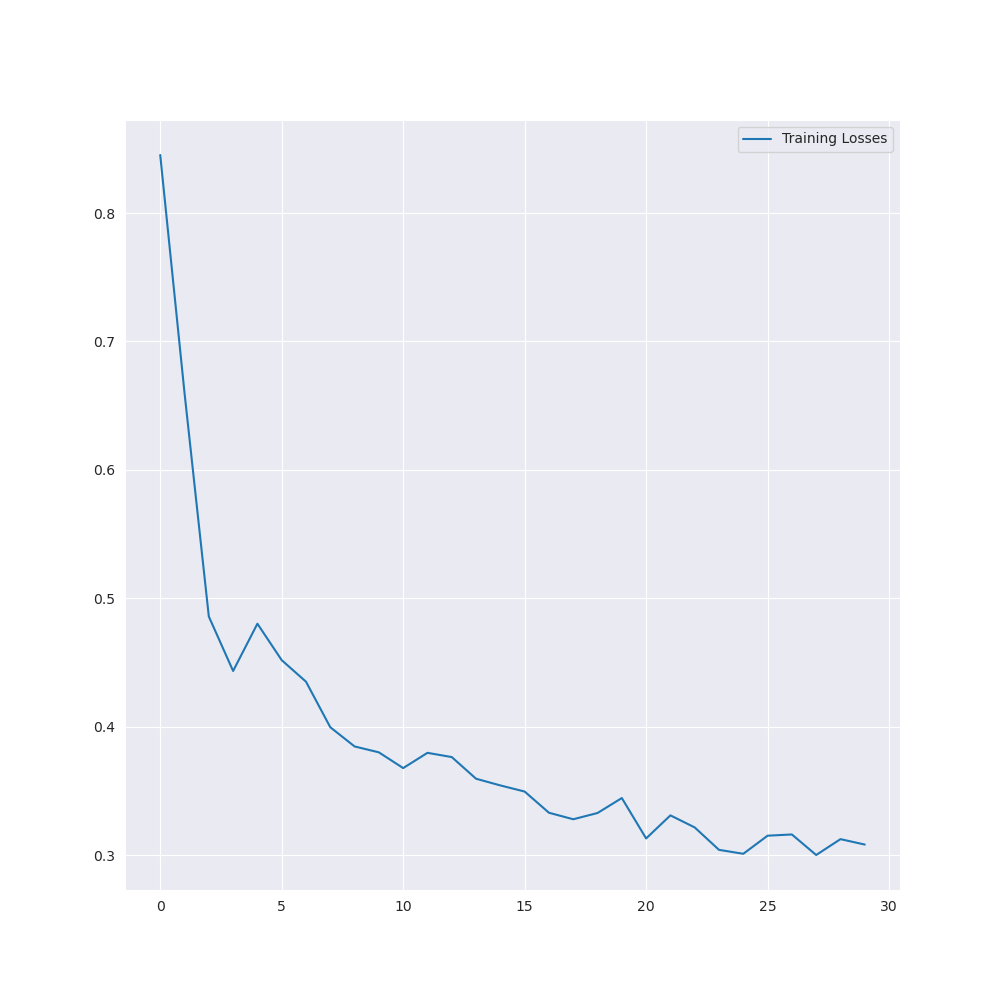}
    	\caption{}
        \label{conv_simsiam_v1}
    \end{subfigure}
    \caption{(a) Convergence behaviour of SimSiam\_v0: Validation of the fact that SimSiam actually converges (while avoiding model collapse) with an arbitrary data augmentation. (b) Convergence behaviour of SimSiam\_v1: Considering standard augmentations with random crops leads to a better convergence. }
    % \label{}
\vspace{-0.4cm}%Put here to reduce too much white space after your table
\end{figure}

% Please add the following required packages to your document preamble:
% \usepackage{multirow}
% \usepackage{graphicx}
\begin{table}[t]
\centering
\resizebox{0.9\linewidth}{!}{%
\begin{tabular}{|c|c|cc|cc|}
\hline
Dataset                 & Metric & \begin{tabular}[c]{@{}c@{}}SimSiam\_v0\\ (w/o \\ norm)\end{tabular} & SimSiam\_v0   & \begin{tabular}[c]{@{}c@{}}SimSiam\_v1\\ (w/o \\ norm)\end{tabular} & SimSiam\_v1   \\ \hline
Data 1                  & CGacc  & 0.5                                                                          & 0.5           & \textbf{1}                                                                   & 0.67          \\ \hline
Data 2                  & CGacc  & 0.5                                                                          & \textbf{0.67} & 0                                                                            & \textbf{0.25} \\ \hline
Data 3                  & CGacc  & 0                                                                            & \textbf{0.4}  & 0                                                                            & \textbf{0.33} \\ \hline
\end{tabular}%
}
\caption{Effect of Embedding Normalization.}
\label{simsiam_normalization}
\vspace{-0.4cm}
\end{table}

We now perform a systematic study of the typical aspects associated with SSL, especially for our particular task of color variants identification. For this purpose, we make use of a single Table \ref{results_all}, where we provide the comparison of various SSL methods, including ours.

\textbf{Convergence behavior with data augmentation for our task:} For illustrating how the convergence behavior of SSL methods changes with respect to a different data augmentation, we pick the SimSiam SSL method for its simplicity and strong claims. We consider two variants of SimSiam: i) SimSiam\_v0: A version of SimSiam, where we used the entire original image as the query, and a color jittered image as the positive, and with a batch size of 12, and ii) SimSiam\_v1: A version of SimSiam with standard SSL \cite{simclr_20} augmentations (\texttt{ColorJitter}, \texttt{RandomGrayscale}, \texttt{RandomHorizontalFlip}, \texttt{GaussianBlur} and \texttt{RandomResizedCrop}), and a batch size of 12. For all cases, the following architecture has been used for SimSiam: \texttt{Encoder}\{ ResNet34 $\rightarrow$ (avgpool) $\rightarrow$ \texttt{ProjectorMLP}(512$\rightarrow$4096$\rightarrow$123) \}$\rightarrow$\texttt{PredictorMLP}(123$\rightarrow$4096$\rightarrow$123).

Figure \ref{conv_simsiam_v0} shows the convergence behaviour of SimSiam\_v0. It can be seen that SimSiam actually converges with an arbitrary data augmentation, while avoiding a model collapse even without making use of negative pairs. Figure \ref{conv_simsiam_v1} shows the convergence behaviour of SimSiam\_v1. We observed that considering standard augmentations with random crops leads to a better convergence than that of SimSiam\_v0. Thus, for all other self-supervised baselines, i.e., BYOL and MOCOv2 we make use of standard augmentations with random crops. However, we later show that our proposed way of considering multiple patches in PBCNet leads to a better performance.%, due to the limitations of the random crops for our use-case.

\textbf{Effect of l2 normalization for our task:} Table \ref{simsiam_normalization} shows the effect of performing $l2$ normalization on the embeddings obtained using SimSiam. We found that without using any normalization, in some cases (eg, Data 2-3) there are no true color variant groups out of the detected clusters (i.e., zero precision), and hence the performance metrics become zero. Thus, for all our later experiments, we make use of $l2$ normalization on the embeddings as a de facto standard, for all the methods.
% Please add the following required packages to your document preamble:
% \usepackage{multirow}
% \usepackage{graphicx}
% \usepackage[table,xcdraw]{xcolor}
% If you use beamer only pass "xcolor=table" option, i.e. \documentclass[xcolor=table]{beamer}
\begin{table*}[t]
\centering
\resizebox{0.7\linewidth}{!}{%
\begin{tabular}{|c|c|c|ccccc|c|}
\hline
\multicolumn{2}{|c|}{}            & {\color[HTML]{3531FF} \textbf{Supervised}}  & \multicolumn{6}{c|}{Self-Supervised}                                                    \\ \hline
Dataset                  & Metric & {\color[HTML]{3531FF} \textbf{Triplet Net}} & SimSiam\_v0 & SimSiam\_v1 & SimSiam\_v2 & BYOL  & MOCOv2       & \textbf{PBCNet (Ours)} \\ \hline
Data 1                   & CGacc  & {\color[HTML]{3531FF} \textbf{0.67}}        & 0.5         & 0.67        & 1           & 0.5   & 1            & \textbf{1}             \\ \hline
Data 2                   & CGacc  & {\color[HTML]{3531FF} \textbf{1}}           & 0.67        & 0.25        & 1           & 0.75  & \textbf{0.8} & 0.75                   \\ \hline
Data 3                   & CGacc  & {\color[HTML]{3531FF} \textbf{0.75}}        & 0.4         & 0.33        & 0           & 0.5   & 0.5          & \textbf{0.6}           \\ \hline
                         & CGacc  & {\color[HTML]{3531FF} \textbf{0.67}}        & 0.4         & 0.5         & 0.5         & 0.5   & \textbf{1}   & 0.85                   \\
                         & ARI    & {\color[HTML]{3531FF} \textbf{0.69}}        & 0.09        & 0.15        & 0.12        & 0.27  & 0.66         & \textbf{0.75}          \\
                         & FMS    & {\color[HTML]{3531FF} \textbf{0.71}}        & 0.15        & 0.22        & 0.20        & 0.30  & 0.71         & \textbf{0.76}          \\
\multirow{-4}{*}{Data 4} & CScore & {\color[HTML]{3531FF} \textbf{0.700}}       & 0.110       & 0.182       & 0.152       & 0.281 & 0.680        & \textbf{0.756}         \\ \hline
                         & CGacc  & {\color[HTML]{3531FF} \textbf{1}}           & 0           & 0.5         & 0.33        & 0.5   & 1            & \textbf{1}             \\
                         & ARI    & {\color[HTML]{3531FF} \textbf{1}}           & 0           & 0.09        & 0.28        & 0.64  & 1            & \textbf{1}             \\
                         & FMS    & {\color[HTML]{3531FF} \textbf{1}}           & 0.22        & 0.30        & 0.45        & 0.71  & 1            & \textbf{1}             \\
\multirow{-4}{*}{Data 5} & CScore & {\color[HTML]{3531FF} \textbf{1}}           & 0           & 0.135       & 0.341       & 0.674 & 1            & \textbf{1}             \\ \hline
                         & CGacc  & {\color[HTML]{3531FF} \textbf{0.83}}        & 0.5         & 0.5         & 0.8         & 0.6   & 1            & \textbf{1}             \\
                         & ARI    & {\color[HTML]{3531FF} \textbf{0.44}}        & 0.07        & 0.06        & 0.04        & 0.20  & 0.58         & \textbf{0.79}          \\
                         & FMS    & {\color[HTML]{3531FF} \textbf{0.49}}        & 0.12        & 0.17        & 0.13        & 0.24  & 0.64         & \textbf{0.80}          \\
\multirow{-4}{*}{Data 6} & CScore & {\color[HTML]{3531FF} \textbf{0.466}}       & 0.089       & 0.090       & 0.063       & 0.214 & 0.610        & \textbf{0.796}         \\ \hline
\end{tabular}%
}
\caption{Comparison of our proposed method against the supervised and state-of-the-art self-supervised baselines, across all the datasets.}
\label{results_all}
\vspace{-0.6cm}
\end{table*}

\textbf{Effect of Batch Size and Momentum Encoding in SSL for our task:}
% Table \ref{effect_batch_sz} shows the effect of batch size on the performance in our task using self-supervised learning. 
For studying the effect of batch size in SSL for our task, we introduce a third variant of SimSiam, i.e., SimSiam\_v2: This is essentially SimSiam\_v1 with a batch size of 128. We then consider SimSiam\_v1, SimSiam\_v2 and BYOL, where the first makes use of a batch size of 12, while the others make use of batch sizes of 128. We observed that a larger batch size usually leads to a better performance. This is observed from Table \ref{results_all}, by the higher values of performance metrics in the columns for SimSiam\_v2 and BYOL (vs SimSiam\_v1). Additionally, we noted that the momentum encoder used in BYOL causes a further boost in the performance, as observed in its superior performance as compared to SimSiam\_v2 that has the same batch size. It should be noted that except for the momentum encoder, the rest of the architecture and augmentations used in BYOL are exactly the same as in SimSiam. We observed that increasing the batch size in SimSiam does not drastically or consistently improve its performance, something which its authors also noticed \cite{simsiam_21}.

\textbf{Effect of Memory Queue in SSL for our task:} %Using Table \ref{effect_queue}, 
We also inspect the effect of an extra memory module/queue being used to facilitate the comparisons with a large number of negative examples. In particular, we make use of the MOCOv2 method with the following settings: i) queue size of 5k, ii) temperature parameter of 0.05, iii) a MLP (512$\rightarrow$4096$\rightarrow$relu$\rightarrow$123) added after the avgpool layer of the ResNet34, iv) SGD for updating the query encoder, with learning rate of 0.001, momentum of 0.9, weight decay of $1e^{-6}$, and v) value of 0.999 for $\theta$ in the momentum update. It is observed from Table \ref{results_all}, by the columns of MOCOv2 and BYOL, that the performance of the former is superior. As BYOL does not use a memory module, but MOCOv2 does, we conclude that using a separate memory module significantly boosts the performance of SSL in our task. Motivated by our observations so far, we choose to employ both momentum encoding and memory module in our proposed PBCNet method.

\subsection{Comparison of PBCNet against the state-of-the-art}
In Table \ref{results_all}, we provide the comparison of our proposed self-supervised method \textbf{PBCNet} against various self-supervised state-of-the-art baselines and the supervised baseline across all the datasets. It should be noted that in Table \ref{results_all}, any performance gains for a specific method is due to the intrinsic nature of the same, and not because of a particular hyperparameter setting. This is because we report the \textit{best performance for each method} after adequate tuning of distance threshold (details in supplementary) and other parameters, and not just their default hyperparameters.
%Also, for each method, the distance threshold used to obtain these results, and the corresponding number of \textit{detected} as well as the number of \textit{correct} groups obtained, are reported in Table \ref{config_all}. From Table \ref{config_all}, we observed that our method is capable of detecting more number of color groups with a usually higher precision, when compared to its competitors. 
Following are the configurations that we have used in our \textbf{PBCNet} method: i) memory module size of 5k, ii) temperature parameter of 0.05, iii) the FC layer after the avgpool layer of the ResNet34 was removed, iv) SGD for updating the query encoder, with learning rate of 0.001, momentum of 0.9, weight decay of $1e^{-6}$, and v) value of 0.999 for $\theta$ in the momentum update.

We made use of a batch size of $32 (=128/4)$ as we have to store tensors for each of the 4 slices simultaneously for each mini-batch (we used a batch size of 128 for the other methods). For data augmentation, we first apply a color distortion in the following order: i) \texttt{ColorJitter(0.8 * s, 0.8 * s, 0.8 * s, 0.2 * s)} with s=1, p=0.8, ii) \texttt{RandomGrayscale} with p=0.2, iii) \texttt{GaussianBlur((3, 3), (1.0, 2.0))} with p=0.5. After the color distortion, we apply our slicing technique. For the second image (positive/negative) we apply the same series of transformations.
\begin{figure}[t]
  \centering
  \includegraphics[width=0.8\linewidth]{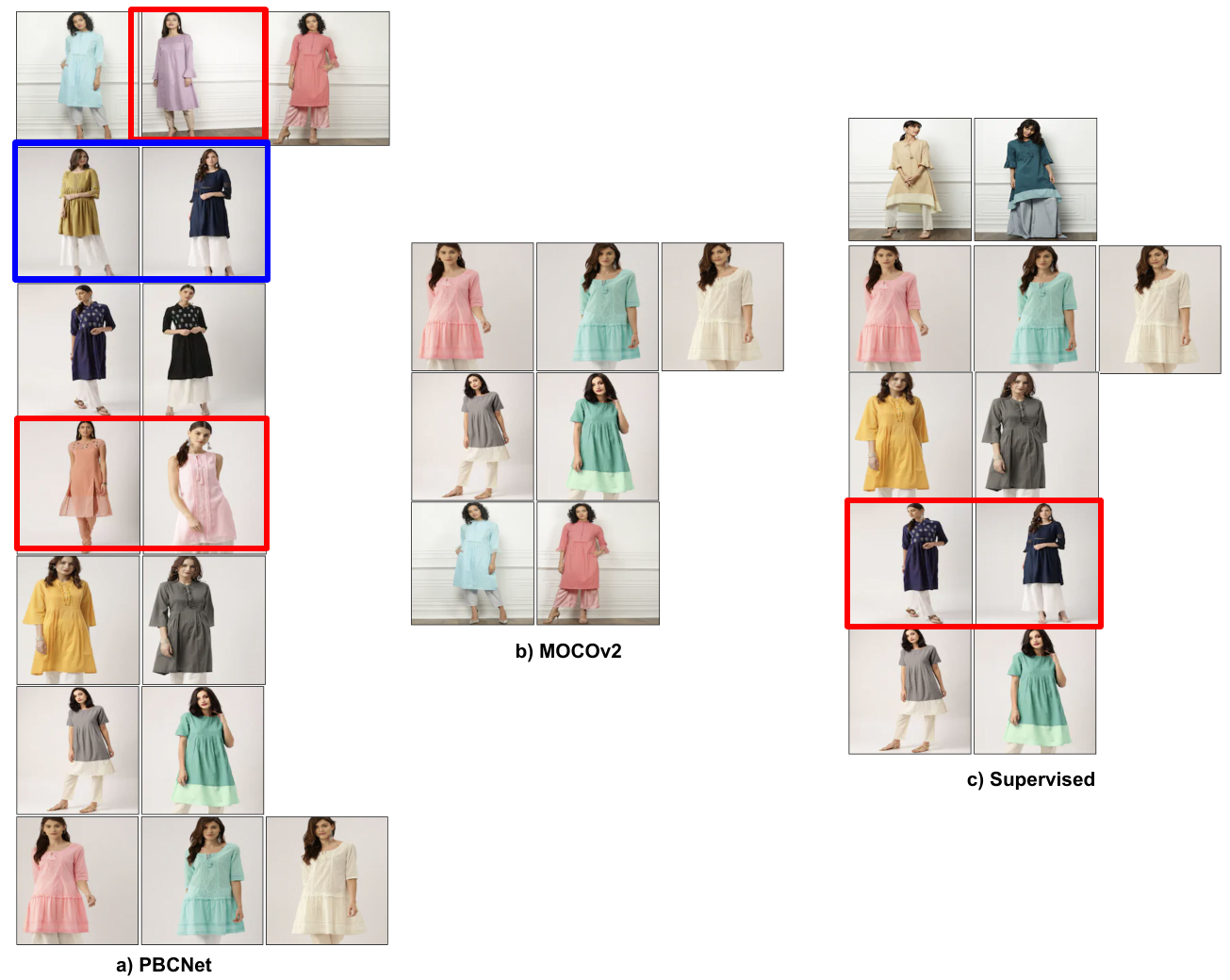}
  \caption{Qualitative comparison of color variants groups obtained using our PBCNet method (left column), MOCOv2 (middle column) and the supervised baseline (right column), on \textit{Data 4}.}
  \label{Data4_quali}
\vspace{-0.6cm}
\end{figure}

From Table \ref{results_all}, it is clear that the SimSiam method despite its strong claims of not using any negative pairs, nor momentum encoder nor large batches, performs poorly as compared to our supervised method (shown in bold blue color). The BYOL method which also do not make use of negative pairs, performs better than the SimSiam method in our use case, by virtue of its momentum encoder.

Among all compared SSL baselines, it is the MOCOv2 method that performs the best. This is due to the reason of the memory queue that facilitates the comparison with a large number of negative examples. This shows that the importance of considering negative pairs still holds true, especially for challenging use-cases like the one considered in the paper. However, our proposed self-supervised method \textbf{PBCNet} clearly outperforms all the baselines. The fact that it outperforms MOCOv2 can be attributed to the patch-based slicing used, which is the only different component in our method in comparison to MOCOv2 that uses random crop. Another interesting thing that we observed is the fact that despite using much lesser batch size of 32, our method outperforms the baselines. In a way, we were able to extract and leverage more information by virtue of the slicing (by borrowing information from the other patches simultaneously), even with smaller batches.

We also noticed that the supervised baseline performs quite good in our task, even without any data augmentation pipeline as used in the SSL methods. However, by virtue of data augmentations like color jitter and cropping, which are pretty relevant to the task of color variants identification, stronger SSL methods like MOCOv2 and PBCNet are in fact capable of surpassing the performance of the supervised baseline as well, in some cases. Having said that, if we do not have adequate labeled data in the first place, we cannot even use supervised learning. Hence, enabling data augmentations and slicing strategy in the supervised model has not been focused, because the necessity of our approach comes from the issue of addressing the lack of labeled data, and not to improve the performance of supervised learning (which any how is label dependent).

\textbf{Effect of Clustering:}
% In Table \ref{effect_agglo}, we report the performance of all the compared self-supervised methods (including ours) on all the datasets, with respect to the different metrics against varying values of the distance threshold used in the Agglomerative clustering. 
In Table \ref{effect_cluster}, we report the performances obtained by varying the clustering algorithm to group embeddings obtained by different SSL methods, on \textit{Data 4-6}. We picked the Agglomerative, DBSCAN and Affinity Propagation clustering techniques that do not require the number of clusters as input parameter (which is difficult to obtain in our use-case). In general, we observed that the Agglomerative clustering technique leads to a better performance in our use-case. Also, for a fixed clustering approach, using embeddings obtained by our PBCNet method usually leads to a better performance.% It should be noted that for a method, only those thresholds are reported against a dataset, using which Agglomerative Clustering produces non-zero performance metrics, thus indicating a meaningful clustering.
% Please add the following required packages to your document preamble:
% \usepackage{graphicx}
\begin{table}[t]
\centering
\resizebox{0.7\linewidth}{!}{%
\begin{tabular}{|c|c|cc|cc|cc|}
\hline
\multicolumn{2}{|c|}{Dataset} & \multicolumn{2}{c|}{\textbf{Data 4}} & \multicolumn{2}{c}{\textbf{Data 5}} & \multicolumn{2}{c|}{\textbf{Data 6}} \\ \hline
Method           & Clustering & \textbf{ARI}      & \textbf{FMS}     & \textbf{ARI}     & \textbf{FMS}     & \textbf{ARI}      & \textbf{FMS}     \\ \hline
\textbf{PBCNet}  & Agglo      & \textbf{0.75}     & \textbf{0.76}    & \textbf{1.00}    & \textbf{1.00}    & \textbf{0.79}     & \textbf{0.80}    \\
                 & DBSCAN     & 0.66              & 0.71             & 1.00             & 1.00             & 0.66              & 0.71             \\
                 & Affinity   & 0.30              & 0.42             & 0.22             & 0.41             & 0.24              & 0.37             \\ \hline
\textbf{MOCOv2}  & Agglo      & \textbf{0.66}     & \textbf{0.71}    & \textbf{1.00}    & \textbf{1.00}    & \textbf{0.58}     & \textbf{0.64}    \\
                 & DBSCAN     & 0.66              & 0.71             & 1.00             & 1.00             & 0.37              & 0.40             \\
                 & Affinity   & 0.20              & 0.32             & 0.04             & 0.26             & 0.25              & 0.41             \\ \hline
\textbf{BYOL}    & Agglo      & \textbf{0.27}     & \textbf{0.30}    & \textbf{0.64}    & \textbf{0.71}    & \textbf{0.20}     & \textbf{0.24}    \\
                 & DBSCAN     & 0.17              & 0.28             & 0.64             & 0.71             & 0.02              & 0.24             \\
                 & Affinity   & 0.03              & 0.14             & 0.28             & 0.45             & 0.01              & 0.11             \\ \hline
\end{tabular}%
}
\caption{Effect of the clustering technique used.}
\label{effect_cluster}
\vspace{-0.5cm}
\end{table}

%The distance thresholds for the various methods in Agglomerative clustering used to obtain the results in Table \ref{results_all} across all the datasets have also been provided in Table \ref{config_all}. The number of detected color groups, as well as the number of correct color groups obtained have also been provided for each dataset. We observed that our method is capable of detecting more number of color groups with a usually higher precision, when compared to its competitors.
\begin{figure}[t]
  \centering
  \includegraphics[width=0.75\linewidth]{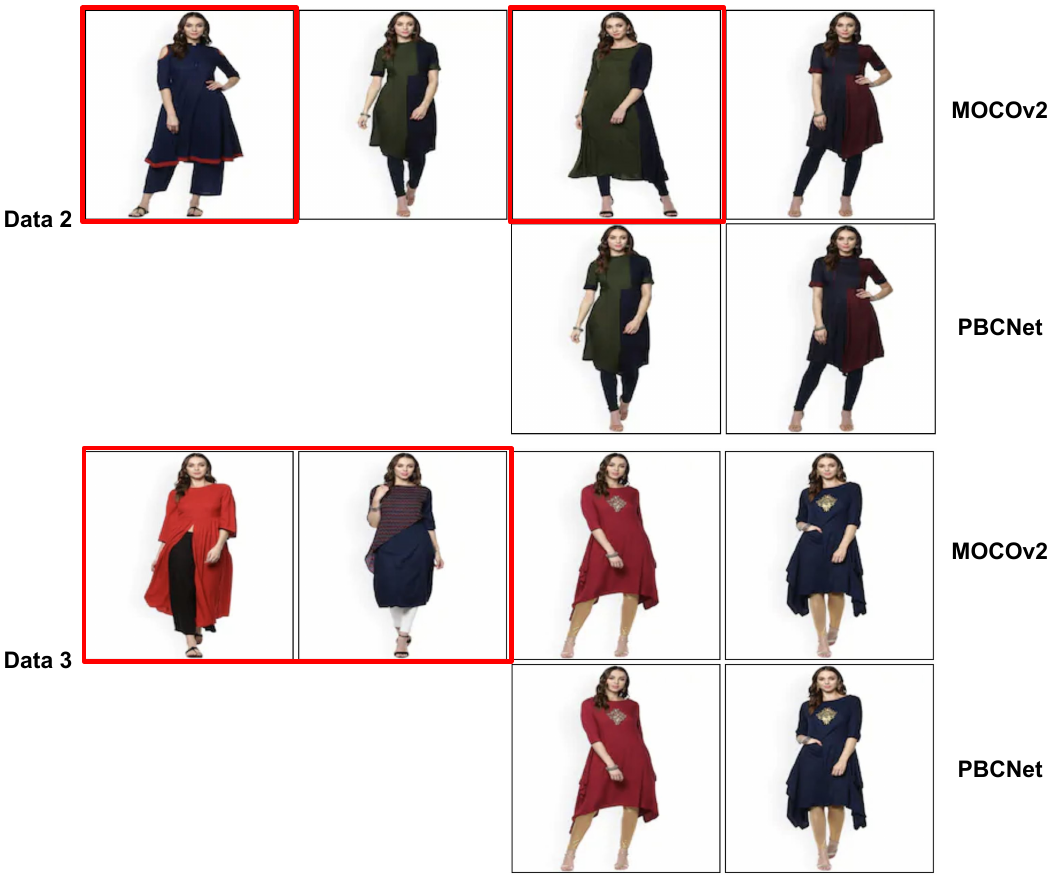}
  \caption{A few groups obtained on \textit{Data 2 \& 3} using MOCOv2 have false positives (shown in red box), while our PBCNet method does not yield such groups.}
  \label{PBCNet_vs_MOCOv2_quali}
\vspace{-0.5cm}
\end{figure}

\begin{figure}[t]
  \centering
  \includegraphics[width=\columnwidth]{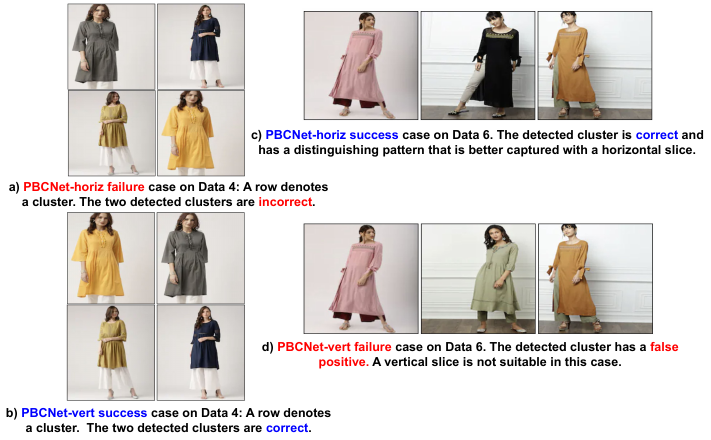}
  \caption{Trade off between vertical and horizontal slicing.}
  \label{slicing_tradeoff}
\vspace{-0.5cm}
\end{figure}

% Please add the following required packages to your document preamble:
% \usepackage{graphicx}
\begin{table}[t]
\centering
\resizebox{\columnwidth}{!}{%
\begin{tabular}{|c|c|c|c|ccc|ccc|ccc|}
\hline
Dataset         & Data 1 & Data 2     & Data 3       & \multicolumn{3}{c|}{Data 4}                                                             & \multicolumn{3}{c|}{Data 5} & \multicolumn{3}{c|}{Data 6}                                   \\ \hline
Method          & CGacc  & CGacc      & CGacc        & CGacc         & ARI                                & FMS                                & CGacc     & ARI    & FMS    & CGacc & ARI                       & FMS                       \\ \hline
PBCNet-horiz    & 1      & \textbf{1} & 0.5          & 0.66          & 0.65                               & 0.67                               & 1         & 1      & 1      & 1     & \textbf{0.81}             & \textbf{0.82}             \\
PBCNet-vert     & 1      & 0.6        & 0.6          & \textbf{1} & \textbf{0.88} & \textbf{0.89} & 1         & 1      & 1      & 0.83  & 0.48 & 0.50 \\ \hline
\textbf{PBCNet} & 1      & 0.75       & \textbf{0.6} & 0.85          & 0.75                               & 0.76          & 1         & 1      & 1      & 1     & 0.79 & 0.80 \\ \hline
\end{tabular}%
}
\caption{Effect of Slicing on PBCNet}
\label{effect_slicing}
\vspace{-0.7cm}
\end{table}
\textbf{Qualitative results:} Sample qualitative comparisons of color variants groups obtained on \textit{Data 4} using our PBCNet method, MOCOv2 and the supervised baseline are provided in Figure \ref{Data4_quali}. Each of the rows for a column corresponding to a method represents a detected color variants cluster for the considered method. A row has been marked with a red box if the entire cluster contains images that are not color variants to each other. A single image is marked with a red box if it is the only incorrect image, while rest of the images are color variants. We observed that our method not only detects clusters with higher precision (which MOCOv2 does as well), but it also has a higher recall, which is comparable to the supervised method. We also make use of a blue box to show a detected color group by our method which contains images that are color variants, but are difficult to be identified at a first glance.

Additionally, Figure \ref{PBCNet_vs_MOCOv2_quali} shows a few color groups identified in the datasets \textit{Data 2 \& 3} using MOCOv2 and our PBCNet. We observed that MOCOv2 detected groups with false positives, while our PBCNet method did not. This could happen because when a random crop is obtained by MOCOv2, it need not necessarily be from a \textit{distinctive} region of an apparel that helps to identify its color variant  (eg, in Figure \ref{PBCNet_vs_MOCOv2_quali}, the bent line like pattern separating the colored and black region of the apparels of Data 2, and the diamond like shape in the apparels of Data 3). We argue that a random crop might have arrived from such a \textit{distinctive} region given that the size of the crop is made larger, etc. But that still leaves things to random chance. On the other hand, our slicing technique being deterministic in nature, \textit{guarantees} that all the regions of an object \textit{would} be captured. We would also like to mention that our slicing approach is agnostic to the fashion apparel type, i.e, the same is easily applicable for any fashion article type (Tops, Shirts, Shoes, Trousers, Track pants, Sweatshirts, etc). In fact, this is how a human identifies color variants as well, by looking at the article along both horizontal and vertical directions, to identify distinctive patterns. Even humans cannot identify an object if we restrict our vision to only a particular small crop.

\textbf{Effect of the slicing}: We also study 2 variants of our PBCNet method: i) PBCNet-horiz (computing an embedding only by considering the top and bottom slices), and ii) PBCNet-vert (computing an embedding using only the left and right slices). The results are shown in Table \ref{effect_slicing}. In \textit{Data 4}, PBCNet-vert performs better than PBCNet-horiz, and in \textit{Data 6}, PBCNet-horiz performs better than PBCNet-vert (significantly). The performance of the two versions is also illustrated in Figure \ref{slicing_tradeoff}. We observed that a single slicing do not work in all scenarios, especially for apparels.

Although the horizontal slicing is quite competitive, it may be beneficial to consider the vertical slices as well. This is observed by the drop in performance of PBCNet-horiz in \textit{Data 3-4} (vs PBCNet). This is because some garments may contain distinguishing patterns that may be better interpreted only by viewing vertically, for example, printed texts (say, \textit{adidas} written vertically), floral patterns etc. In such cases, simply considering horizontal slices may actually split/ disrupt the vertical information. It may also happen that mixing of slicing introduces some form of redundancy, as observed by the occasional drop in the performance of PBCNet when compared to PBCNet-horiz (on \textit{Data 6}) and PBCNet-vert (on \textit{Data 4}). However, on average PBCNet leads to an overall consistent and competitive performance, while avoiding drastically fluctuating improvements or failures. We suggest considering both the directions of slicing, so that they could collectively represent all necessary and distinguishing patterns, and if one slicing misses some important information, the other could compensate for it.
% It should also be noted that we consider both horizontal (yields top and bottom) and vertical (yields left and right) slicing simultaneously. This is because apparels most often contain printed texts and floral patterns, which may be present in either horizontal or vertical format. For example, consider a garment with the text \textit{adidas} written in a vertical manner. In this case, considering only horizontal slicing would yield a slice that splits the text, and hence the vertical slice would be more suitable. Similarly, for a pattern/text that is horizontal in nature, splitting vertically would disrupt the content in a slice. However, detecting each and every pattern for garments is a difficult task. Thus, to address this, we consider all the possible directions of slicing, with the hope that they would collectively represent all necessary and distinguishing patterns.

\section{Conclusions}
In this paper, we utilize deep visual Representation Learning to address the problem of identification of color variants (images of objects exactly similar in design, but not color), particularly for fashion products. A supervised triplet loss based deep neural network model for visual Representation Learning has been proposed to identify the color variants. A systematic study of existing state-of-the-art self-supervised methods has been done to solve the proposed problem, while alleviating the need for manual annotations. Also, a novel contrastive loss based self-supervised representation learning method that focuses on parts of an object has been proposed. This is done to make the model better informed of the discriminative regions of an image, in order to identify color variants.

% \subssonubsection*{Acknowledgements}
% We would like to thank the anonymous reviewers for their efforts at reviewing our work.

%\clearpage
{\small
\bibliographystyle{named}
\bibliography{PBCNet_IJCAI21}
}

% \pagebreak

\section{Supplementary: Additional Details and Results (omitted from the main text due to space constraints)}
We evaluated the discussed methods on a large (orders of magnitudes of $10^5$) internal collection of challenging real-world industrial images on our Myntra platform (\url{www.myntra.com}) that hosts various fashion products. In this section, we report our results on a collection of roughly\footnote{Company compliance policies prohibit open-sourcing/ revealing exact dataset specifics} 0.13 million Kurtas images from our internal database. A disjoint set of 6k images were manually annotated by our in-house team to provide class labels indicating their color variants information. This labeled data was used to form triplets to train the supervised triplet loss based color variants model. We also used the exact same set to train the self-supervised methods for a fair comparison. For inferencing, we used the entire 0.13 million Kurtas images, which are present in the form of different dataset splits (based on brand, gender, MRP). We refer to our 6 dataset splits as Data 1-6. Table \ref{dataset_details} provides details of the datasets used to compare the methods, along with their meta-data on our platform. It should be noted that the datasets \textit{Data 1-3} do not have manual annotations available for them, whereas the remaining three have been annotated by the taggers.
\begin{table}[!htb]
\centering
\resizebox{0.8\linewidth}{!}{%
\begin{tabular}{|c|cccc|c|}
\hline
\textbf{Dataset} & \textbf{Article\_Type}        & \textbf{Gender}              & \textbf{Brand}                       & \textbf{MRP}                & \textbf{\begin{tabular}[c]{@{}c@{}}Ground Truth\\ Available\end{tabular}} \\ \hline
Data 1           & Kurtas                        & Women                        & STREET 9                             & 1899                        & No                                                                        \\
Data 2           & Kurtas                        & Women                        & STREET 9                             & 1399                        & No                                                                        \\
Data 3           & Kurtas                        & Women                        & STREET 9                             & 1499                        & No                                                                        \\
Data 4           & Kurtas                        & Women                        & all about you                        & 1699                        & Yes                                                                       \\
Data 5           & Kurtas                        & Women                        & all about you                        & 2499                        & Yes                                                                       \\
Data 6           & Kurtas                        & Women                        & all about you                        & 2199                        & Yes                                                                       \\ \hline
\end{tabular}%
}
\caption{Details of datasets used to compare the methods.}
\label{dataset_details}
\vspace{-0.6cm}
\end{table}

% Please add the following required packages to your document preamble:
% \usepackage{multirow}
% \usepackage{graphicx}
\begin{table}[!htb]
\centering
\resizebox{0.6\linewidth}{!}{%
\begin{tabular}{|c|c|c|ccccc|c|}
\hline
Dataset                 & Config    & Sup & S0 & S1  & S2 & B & M & PB \\ \hline
\multirow{3}{*}{Data 1} & t        & 1          & 0.07        & 0.3          & 0.3         & 0.5  & 0.7    & 0.8    \\ \cline{2-9} 
                        & nCG  & 2          & 1           & 2            & 1           & 1    & 2      & 2      \\
                        & nDG & 3          & 2           & 3            & 1           & 2    & 2      & 2      \\ \hline
\multirow{3}{*}{Data 2} & t        & 0.87       & 0.07        & 0.15         & 0.2         & 0.5  & 0.8    & 0.8    \\ \cline{2-9} 
                        & nCG  & 4          & 2           & 1            & 2           & 3    & 4      & 3      \\
                        & nDG & 4          & 3           & 4            & 2           & 4    & 5      & 4      \\ \hline
\multirow{3}{*}{Data 3} & t        & 0.87       & 0.2         & 0.24 & 0.3  & 0.5  & 0.8    & 0.8    \\ \cline{2-9} 
                        & nCG  & 3          & 2           & 2            & 0           & 1    & 3      & 3      \\
                        & nDG & 4          & 5           & 6            & 5 to 7      & 2    & 6      & 5      \\ \hline
\multirow{3}{*}{Data 4} & t        & 0.8        & 0.1         & 0.1          & 0.5         & 0.5  & 0.6    & 0.8    \\ \cline{2-9} 
                        & nCG  & 4          & 2           & 1            & 3           & 3    & 3      & 6      \\
                        & nDG & 6          & 5           & 2            & 6           & 6    & 3      & 7      \\ \hline
\multirow{3}{*}{Data 5} & t        & 0.9        & 0.3         & 0.6          & 0.6         & 0.7  & 0.7    & 0.8    \\ \cline{2-9} 
                        & nCG  & 1          & 0           & 1            & 1           & 1    & 1      & 1      \\
                        & nDG & 1          & 1           & 2            & 3           & 2    & 1      & 1      \\ \hline
\multirow{3}{*}{Data 6} & t        & 0.95       & 0.07        & 0.8          & 0.8         & 0.6  & 0.8    & 0.9    \\ \cline{2-9} 
                        & nCG  & 5          & 2           & 2            & 4           & 3    & 6      & 7      \\
                        & nDG & 6          & 4           & 4            & 5           & 5    & 6      & 7      \\ \hline
\end{tabular}%
}
\caption{Distance thresholds in Agglomerative clustering used to produce the \textit{best} results (for each method) across all the datasets, along with the number of \textit{detected} and \textit{correct} color groups. NOTE that we have used the following notations: Config: Configuration, t:threshold, nCG:n\_correct\_gps, nDG:n\_detected\_gps, Sup:Supervised, S0:SimSiam\_v0, S1:SimSiam\_v1, S2:SimSiam\_v2, B:BYOL, M:MOCOv2, and PB:PBCNet.}
\label{config_all}
\vspace{-0.4cm}
\end{table}

% Please add the following required packages to your document preamble:
% \usepackage{multirow}
% \usepackage{graphicx}
\begin{table*}[!htb]
\centering
\resizebox{0.7\linewidth}{!}{%
\begin{tabular}{|c|c|c|cc|c|c|c|cc|c|c|c|cc|}
\hline
Method                           & \multicolumn{4}{c|}{\textbf{SimSiam\_v0}}                                           & Method                           & \multicolumn{4}{c|}{\textbf{SimSiam\_v1}}                                           & Method                           & \multicolumn{4}{c|}{\textbf{SimSiam\_v2}}                                           \\ \hline \hline
Dataset                          & \textbf{dist\_threshold} & \textbf{CGacc} & \textbf{ARI}        & \textbf{FMS}      & Dataset                          & \textbf{dist\_threshold} & \textbf{CGacc} & \textbf{ARI}       & \textbf{FMS}       & Dataset                          & \textbf{dist\_threshold} & \textbf{CGacc} & \textbf{ARI}       & \textbf{FMS}       \\ \hline
\multirow{3}{*}{\textbf{Data 1}} & 0.07                     & 0.5            & \multicolumn{2}{c|}{\multirow{9}{*}{}}  & \multirow{4}{*}{\textbf{Data 1}} & 0.2                      & 1              & \multicolumn{2}{c|}{\multirow{13}{*}{}} & \multirow{3}{*}{\textbf{Data 1}} & 0.3                      & 1              & \multicolumn{2}{c|}{\multirow{10}{*}{}} \\
                                 & 0.1                      & 0.33           & \multicolumn{2}{c|}{}                   &                                  & 0.3                      & 0.67           & \multicolumn{2}{c|}{}                   &                                  & 0.5                      & 0.5            & \multicolumn{2}{c|}{}                   \\
                                 & 0.2                      & 0.33           & \multicolumn{2}{c|}{}                   &                                  & 0.4                      & 0.5            & \multicolumn{2}{c|}{}                   &                                  & 0.6                      & 0.5            & \multicolumn{2}{c|}{}                   \\ \cline{1-3} \cline{11-13}
\multirow{3}{*}{\textbf{Data 2}} & 0.07                     & 0.67           & \multicolumn{2}{c|}{}                   &                                  & 0.5                      & 0.67           & \multicolumn{2}{c|}{}                   & \multirow{4}{*}{\textbf{Data 2}} & 0.2                      & 1              & \multicolumn{2}{c|}{}                   \\ \cline{6-8}
                                 & 0.1                      & 0.4            & \multicolumn{2}{c|}{}                   & \multirow{3}{*}{\textbf{Data 2}} & 0.15                     & 0.25           & \multicolumn{2}{c|}{}                   &                                  & 0.3                      & 0.4            & \multicolumn{2}{c|}{}                   \\
                                 & 0.2                      & 0.25           & \multicolumn{2}{c|}{}                   &                                  & 0.2                      & 0.2            & \multicolumn{2}{c|}{}                   &                                  & 0.4                      & 0.5            & \multicolumn{2}{c|}{}                   \\ \cline{1-3}
\multirow{3}{*}{\textbf{Data 3}} & 0.07                     & 0              & \multicolumn{2}{c|}{}                   &                                  & 0.3                      & 0.2            & \multicolumn{2}{c|}{}                   &                                  & 0.5                      & 0.5            & \multicolumn{2}{c|}{}                   \\ \cline{6-8} \cline{11-13}
                                 & 0.1                      & 0              & \multicolumn{2}{c|}{}                   & \multirow{6}{*}{\textbf{Data 3}} & 0.1                      & 0              & \multicolumn{2}{c|}{}                   & \multirow{3}{*}{\textbf{Data 3}} & 0.2                      & 0              & \multicolumn{2}{c|}{}                   \\
                                 & 0.2                      & 0.4            & \multicolumn{2}{c|}{}                   &                                  & 0.2                      & 0.2            & \multicolumn{2}{c|}{}                   &                                  & 0.3                      & 0              & \multicolumn{2}{c|}{}                   \\ \cline{1-5}
\textbf{Data 4}                  & 0.1                      & 0.4            & 0.09                & 0.15              &                                  & 0.23 to 0.25             & 0.33           & \multicolumn{2}{c|}{}                   &                                  & 0.5                      & 0              & \multicolumn{2}{c|}{}                   \\ \cline{1-5} \cline{11-15} 
\textbf{Data 5}                  & 0.3                      & 0              & 0.00                & 0.22              &                                  & 0.3                      & 0.33           & \multicolumn{2}{c|}{}                   & \multirow{2}{*}{\textbf{Data 4}} & 0.3                      & 0.33           & 0.14               & 0.18               \\ \cline{1-5}
\multirow{2}{*}{\textbf{Data 6}} & 0.07                     & 0.5            & 0.07                & 0.12              &                                  & 0.4                      & 0.16           & \multicolumn{2}{c|}{}                   &                                  & 0.5                      & 0.5            & 0.12               & 0.20               \\ \cline{11-15} 
                                 & 0.1                      & 0.4            & 0.01                & 0.08              &                                  & 0.6                      & 0              & \multicolumn{2}{c|}{}                   & \multirow{2}{*}{\textbf{Data 5}} & 0.6                      & 0.33           & 0.28               & 0.45               \\ \cline{1-10}
\multicolumn{5}{|c|}{\multirow{5}{*}{}}                                                                                & \multirow{2}{*}{\textbf{Data 4}} & 0.1                      & 0.5            & 0.15               & 0.22               &                                  & 1.1                      & 0.5            & 0.09               & 0.30               \\ \cline{11-15} 
\multicolumn{5}{|c|}{}                                                                                                 &                                  & 0.2                      & 0.14           & 0.03               & 0.08               & \multirow{3}{*}{\textbf{Data 6}} & 0.6                      & 0.33           & 0.02               & 0.09               \\ \cline{6-10}
\multicolumn{5}{|c|}{}                                                                                                 & \textbf{Data 5}                  & 0.6                      & 0.5            & 0.09               & 0.30               &                                  & 0.8                      & 0.8            & 0.04               & 0.13               \\ \cline{6-10}
\multicolumn{5}{|c|}{}                                                                                                 & \multirow{2}{*}{\textbf{Data 6}} & 0.4                      & 0.4            & 0.05               & 0.14               &                                  & 0.9                      & 0.75           & 0.01               & 0.11               \\ \cline{11-15} 
\multicolumn{5}{|c|}{}                                                                                                 &                                  & 0.8                      & 0.5            & 0.06               & 0.17               & \multicolumn{5}{c|}{}                                                                                                  \\ \hline \hline
Method                           & \multicolumn{4}{c|}{\textbf{BYOL}}                                                  & Method                           & \multicolumn{4}{c|}{\textbf{MOCOv2}}                                                & Method                           & \multicolumn{4}{c|}{\textbf{PBCNet}}                                                \\ \hline \hline
Dataset                          & \textbf{dist\_threshold} & \textbf{CGacc} & \textbf{ARI}        & \textbf{FMS}      & Dataset                          & \textbf{dist\_threshold} & \textbf{CGacc} & \textbf{ARI}       & \textbf{FMS}       & Dataset                          & \textbf{dist\_threshold} & \textbf{CGacc} & \textbf{ARI}       & \textbf{FMS}       \\ \hline
\multirow{3}{*}{\textbf{Data 1}} & 0.5                      & 0.5            & \multicolumn{2}{c|}{\multirow{11}{*}{}} & \multirow{4}{*}{\textbf{Data 1}} & 0.6                      & 1              & \multicolumn{2}{c|}{\multirow{12}{*}{}} & \multirow{5}{*}{\textbf{Data 1}} & 0.6                      & 1              & \multicolumn{2}{c|}{\multirow{14}{*}{}} \\
                                 & 0.7                      & 0.5            & \multicolumn{2}{c|}{}                   &                                  & 0.7                      & 1              & \multicolumn{2}{c|}{}                   &                                  & 0.7                      & 1              & \multicolumn{2}{c|}{}                   \\
                                 & 1.1                      & 0.33           & \multicolumn{2}{c|}{}                   &                                  & 0.8                      & 0.67           & \multicolumn{2}{c|}{}                   &                                  & 0.8                      & 1              & \multicolumn{2}{c|}{}                   \\ \cline{1-3}
\multirow{3}{*}{\textbf{Data 2}} & 0.5                      & 0.75           & \multicolumn{2}{c|}{}                   &                                  & 0.9                      & 0.5            & \multicolumn{2}{c|}{}                   &                                  & 0.9                      & 0.67           & \multicolumn{2}{c|}{}                   \\ \cline{6-8}
                                 & 0.7                      & 0.6            & \multicolumn{2}{c|}{}                   & \multirow{4}{*}{\textbf{Data 2}} & 0.6                      & 0              & \multicolumn{2}{c|}{}                   &                                  & 1                        & 0.5            & \multicolumn{2}{c|}{}                   \\ \cline{11-13}
                                 & 0.9                      & 0.6            & \multicolumn{2}{c|}{}                   &                                  & 0.7                      & 1              & \multicolumn{2}{c|}{}                   & \multirow{4}{*}{\textbf{Data 2}} & 0.6                      & 1              & \multicolumn{2}{c|}{}                   \\ \cline{1-3}
\multirow{5}{*}{\textbf{Data 3}} & 0.5                      & 0.5            & \multicolumn{2}{c|}{}                   &                                  & 0.8                      & 0.8            & \multicolumn{2}{c|}{}                   &                                  & 0.8                      & 0.75           & \multicolumn{2}{c|}{}                   \\
                                 & 0.55                     & 0.33           & \multicolumn{2}{c|}{}                   &                                  & 0.9                      & 0.67           & \multicolumn{2}{c|}{}                   &                                  & 0.9                      & 0.75           & \multicolumn{2}{c|}{}                   \\ \cline{6-8}
                                 & 0.6                      & 0.25           & \multicolumn{2}{c|}{}                   & \multirow{4}{*}{\textbf{Data 3}} & 0.6                      & 0.5            & \multicolumn{2}{c|}{}                   &                                  & 1                        & 0.67           & \multicolumn{2}{c|}{}                   \\ \cline{11-13}
                                 & 0.7                      & 0.33           & \multicolumn{2}{c|}{}                   &                                  & 0.7                      & 0.5            & \multicolumn{2}{c|}{}                   & \multirow{5}{*}{\textbf{Data 3}} & 0.6                      & 0.5            & \multicolumn{2}{c|}{}                   \\
                                 & 0.9                      & 0.28           & \multicolumn{2}{c|}{}                   &                                  & 0.8                      & 0.5            & \multicolumn{2}{c|}{}                   &                                  & 0.7                      & 0.33           & \multicolumn{2}{c|}{}                   \\ \cline{1-5}
\multirow{3}{*}{\textbf{Data 4}} & 0.3                      & 0.5            & 0.15                & 0.22              &                                  & 0.9                      & 0.42           & \multicolumn{2}{c|}{}                   &                                  & 0.8                      & 0.6            & \multicolumn{2}{c|}{}                   \\ \cline{6-10}
                                 & 0.45                     & 0.33           & 0.22                & 0.26              & \multirow{5}{*}{\textbf{Data 4}} & 0.5                      & 1              & 0.17               & 0.32               &                                  & 0.9                      & 0.42           & \multicolumn{2}{c|}{}                   \\
                                 & 0.5                      & 0.5            & 0.27                & 0.30              &                                  & 0.6                      & 1              & 0.66               & 0.71               &                                  & 1                        & 0.42           & \multicolumn{2}{c|}{}                   \\ \cline{1-5} \cline{11-15} 
\multirow{2}{*}{\textbf{Data 5}} & 0.7                      & 0.5            & 0.64                & 0.71              &                                  & 0.7                      & 0.57           & 0.49               & 0.53               & \multirow{5}{*}{\textbf{Data 4}} & 0.5                      & 1              & 0.32               & 0.45               \\
                                 & 0.8                      & 0.33           & 0.46                & 0.58              &                                  & 0.8                      & 0.67           & 0.37               & 0.47               &                                  & 0.6                      & 1              & 0.45               & 0.55               \\ \cline{1-5}
\multirow{2}{*}{\textbf{Data 6}} & 0.55                     & 0.67           & 0.12                & 0.16              &                                  & 0.9                      & 0.6            & 0.29               & 0.42               &                                  & 0.7                      & 1              & 0.74               & 0.77               \\ \cline{6-10}
                                 & 0.6                      & 0.6            & 0.20                & 0.24              & \multirow{3}{*}{\textbf{Data 5}} & 0.7                      & 1              & 1.00               & 1.00               &                                  & 0.8                      & 0.85           & 0.75               & 0.76               \\ \cline{1-5}
\multicolumn{5}{|c|}{\multirow{10}{*}{}}                                                                               &                                  & 0.8                      & 0.5            & 0.64               & 0.71               &                                  & 0.9                      & 0.85           & 0.65               & 0.68               \\ \cline{11-15} 
\multicolumn{5}{|c|}{}                                                                                                 &                                  & 0.9                      & 0.33           & 0.28               & 0.45               & \multirow{4}{*}{\textbf{Data 5}} & 0.6                      & 1              & 1.00               & 1.00               \\ \cline{6-10}
\multicolumn{5}{|c|}{}                                                                                                 & \multirow{8}{*}{\textbf{Data 6}} & 0.6                      & 1              & 0.39               & 0.50               &                                  & 0.7                      & 1              & 1.00               & 1.00               \\
\multicolumn{5}{|c|}{}                                                                                                 &                                  & 0.7                      & 0.67           & 0.34               & 0.37               &                                  & 0.8                      & 1              & 1.00               & 1.00               \\
\multicolumn{5}{|c|}{}                                                                                                 &                                  & 0.8                      & 1              & 0.58               & 0.64               &                                  & 0.9                      & 0.5            & 0.64               & 0.71               \\ \cline{11-15} 
\multicolumn{5}{|c|}{}                                                                                                 &                                  & 0.9                      & 0.85           & 0.56               & 0.65               & \multirow{5}{*}{\textbf{Data 6}} & 0.6                      & 1              & 0.39               & 0.50               \\ \cline{7-10}
\multicolumn{5}{|c|}{}                                                                                                 &                                  & \multicolumn{4}{c|}{\multirow{4}{*}{}}                                              &                                  & 0.7                      & 1              & 0.66               & 0.71               \\
\multicolumn{5}{|c|}{}                                                                                                 &                                  & \multicolumn{4}{c|}{}                                                               &                                  & 0.8                      & 1              & 0.70               & 0.72               \\
\multicolumn{5}{|c|}{}                                                                                                 &                                  & \multicolumn{4}{c|}{}                                                               &                                  & 0.9                      & 1              & 0.79               & 0.80               \\
\multicolumn{5}{|c|}{}                                                                                                 &                                  & \multicolumn{4}{c|}{}                                                               &                                  & 1                        & 1              & 0.47               & 0.53               \\ \hline
\end{tabular}%
}
\caption{Varying values of performance metrics against distance thresholds. For a method, only those thresholds are reported against a dataset, using which Agglomerative Clustering produces non-zero performance metrics, thus indicating a meaningful clustering.}%Varying values of performance metrics on different datasets for the compared self-supervised models with respect to different distance thresholds of the agglomerative clustering. For a dataset, only those thresholds are reported for a method, using which the clustering produces non-zero performance metrics, thus indicating a meaningful clustering.
\label{effect_agglo}
\vspace{-0.45cm}
\end{table*}

\textbf{Performance metrics used:} To evaluate the methods, we made use of the following performance metrics (CGacc and CScore are defined by us for our use-case):
\begin{enumerate}[noitemsep,nolistsep]
    \item Color Group Accuracy (\textbf{CGacc}): It refers to the ratio of the number of \textit{correct color groups} to that of the number of \textit{detected color groups}. Here, \textit{detected color groups} are the clusters identified by the clustering algorithm that have a size of at least two. A \textit{correct color group} is a cluster among the \textit{detected color groups}, which contains at least half of its examples which are actually color variants to each other (while implementing we take a floor function). It should be noted that this performance metric has a direct business relevance due to the fact that it reflects the \textit{precision}. Also, it is computed by our in-house catalog team by performing manual Quality Check (QC).
    \item Adjusted Random Index (\textbf{ARI}): Let s (and d, respectively) be the number of pairs of elements that are in the same (and different, respectively) set in the ground truth class assignment, and in the same (and different, respectively) set in the clustering. Then, the (unadjusted) Random Index is computed as: $RI= \frac{s+d}{^{N}\textrm{C}_{2}}$, where $N$ is the number of elements clustered. The Adjusted RI (ARI) is then computed as: $ARI= \frac{RI-\mathbb{E}[RI]}{\textrm{max}(RI)-\mathbb{E}[RI]}$. ARI ensures that a random label assignment will get a value close to zero (which RI does not).
    \item Fowlkes-Mallows Score (\textbf{FMS}): It is computed as: $FMS=\frac{TP}{\sqrt[]{(TP+FP)(TP+FN)}}$, where $TP$ is the number of pairs that belong to the same clusters in both the ground truth, as well as the predicted cluster labels, $FP$ is the number of pairs that belong to the same clusters in the ground truth, but not in the predicted cluster labels, and $FN$ is the number of pairs that belong in the same clusters in the predicted cluster labels, but not in the ground truth.
    \item Clustering Score (\textbf{CScore}): It is computed as: $CScore = \frac{2.ARI.FMS}{ARI+FMS}$.
\end{enumerate}
It should be noted that while we use CGacc to compare the methods for all the datasets (\textit{Data 1-6}), the remaining metrics are reported only for the datasets \textit{Data 4-6}, where we have the ground-truth labels. Also, all the performance metrics take values in the range $[0,1]$, where a higher value indicates a better performance.

Also, for each method, the distance threshold used to obtain the results, and the corresponding number of \textit{detected} as well as the number of \textit{correct} groups obtained, are reported in Table \ref{config_all}. From Table \ref{config_all}, we observed that our method is capable of detecting more number of color groups with a usually higher precision, when compared to its competitors.

In Table \ref{effect_agglo}, we report the performance of all the compared self-supervised methods (including ours) on all the datasets, with respect to the different metrics against varying values of the distance threshold used in the Agglomerative clustering.

\end{document}